%% file: main.tex
\documentclass[10pt,twocolumn,letterpaper]{article}

\usepackage{cvpr} % uncomment this for the final submission
\usepackage{times}
\usepackage{epsfig}
\usepackage{graphicx}
\usepackage{amsmath}
\usepackage{amssymb}

\usepackage{multirow}
\usepackage{array}
\usepackage{caption}
\usepackage{booktabs}
\usepackage{makecell}

\usepackage[pagebackref,breaklinks,colorlinks,bookmarks=true]{hyperref}

\begin{document}

%%%%%%%%% TITLE
\title{CGTGait: Collaborative Graph and Transformer for Gait Emotion Recognition}

\author{Junjie Zhou$^2$, Haijun Xiong$^1$, Junhao Lu$^3$, Ziyu Lin$^4$, Bin Feng$^{1}$\thanks{Corresponding Author.}\\
$^1$Huazhong University of Science and Technology, China \quad  $^2$Wuhan University of Technology, China\\
$^3$Hefei University of Technology, China \quad $^4$Boston University, United States\\
{\tt\small \{xionghj,fengbin\}@hust.edu.cn, zhoujunjie@whut.edu.cn,}\\
{\tt\small 2021214872@mail.hfut.edu.cn, ziyulin@bu.edu}
% For a paper whose authors are all at the same institution,
% omit the following lines up until the closing ``}''.
% Additional authors and addresses can be added with ``\and'',
% just like the second author.
% To save space, use either the email address or home page, not both
% \and
% Haijun Xiong\\
% Huazhong University of Science and Technology\\
% Wuhan, China\\
% {\tt\small xionghj@hust.edu.cn}
}

\maketitle
\thispagestyle{empty}

%%%%%%%%% ABSTRACT
\input{sec/0_abstract}
%%%%%%%%% BODY TEXT
\input{sec/1_intro}
\input{sec/2_relat}
\input{sec/3_method}
\input{sec/4_exper}
\input{sec/5_conclu}

{\small
\bibliographystyle{ieee}
\bibliography{egbib}
}

\end{document}

%% file: sec/0_abstract.tex
\begin{abstract}
Skeleton-based gait emotion recognition has received significant attention due to its wide-ranging applications. However, existing methods primarily focus on extracting spatial and local temporal motion information, failing to capture long-range temporal representations. In this paper, we propose \textbf{CGTGait}, a novel framework that collaboratively integrates graph convolution and transformers to extract discriminative spatiotemporal features for gait emotion recognition. Specifically, CGTGait consists of multiple CGT blocks, where each block employs graph convolution to capture frame-level spatial topology and the transformer to model global temporal dependencies. Additionally, we introduce a Bidirectional Cross-Stream Fusion (BCSF) module to effectively aggregate posture and motion spatiotemporal features, facilitating the exchange of complementary information between the two streams. We evaluate our method on two widely used datasets, Emotion-Gait and ELMD, demonstrating that our CGTGait achieves state-of-the-art or at least competitive performance while reducing computational complexity by approximately \textbf{82.2\%} (only requiring 0.34G FLOPs) during testing. Code is available at \small\url{https://github.com/githubzjj1/CGTGait}.
\end{abstract}

%% file: sec/1_intro.tex
\section{Introduction}
\label{sec: intro}
Emotion recognition has been widely applied in various fields, including human-computer interaction, video surveillance, and smart healthcare~\cite{wortman2023hicem, khosravi2023crowd, zhao2022fine}. Some studies~\cite{deligianni2019emotions, halovic2018not, li2016identifying, lima2024st} have shown that gait patterns (\eg, walking speed, stride length, and arm swing angles) can reflect an individual's emotional state. Compared to facial emotion recognition~\cite{li2020deep}, gait-based methods do not require close interaction or active cooperation from subjects~\cite{xu2022emotion, li2021static}. With the use of 3D skeletons, gait emotion recognition (GER)~\cite{chen2023ast, lu2023see, zhai2024looking} has gained significant attention in the computer vision community. 

\begin{figure}
    \centering
    \includegraphics[width=0.95\linewidth]{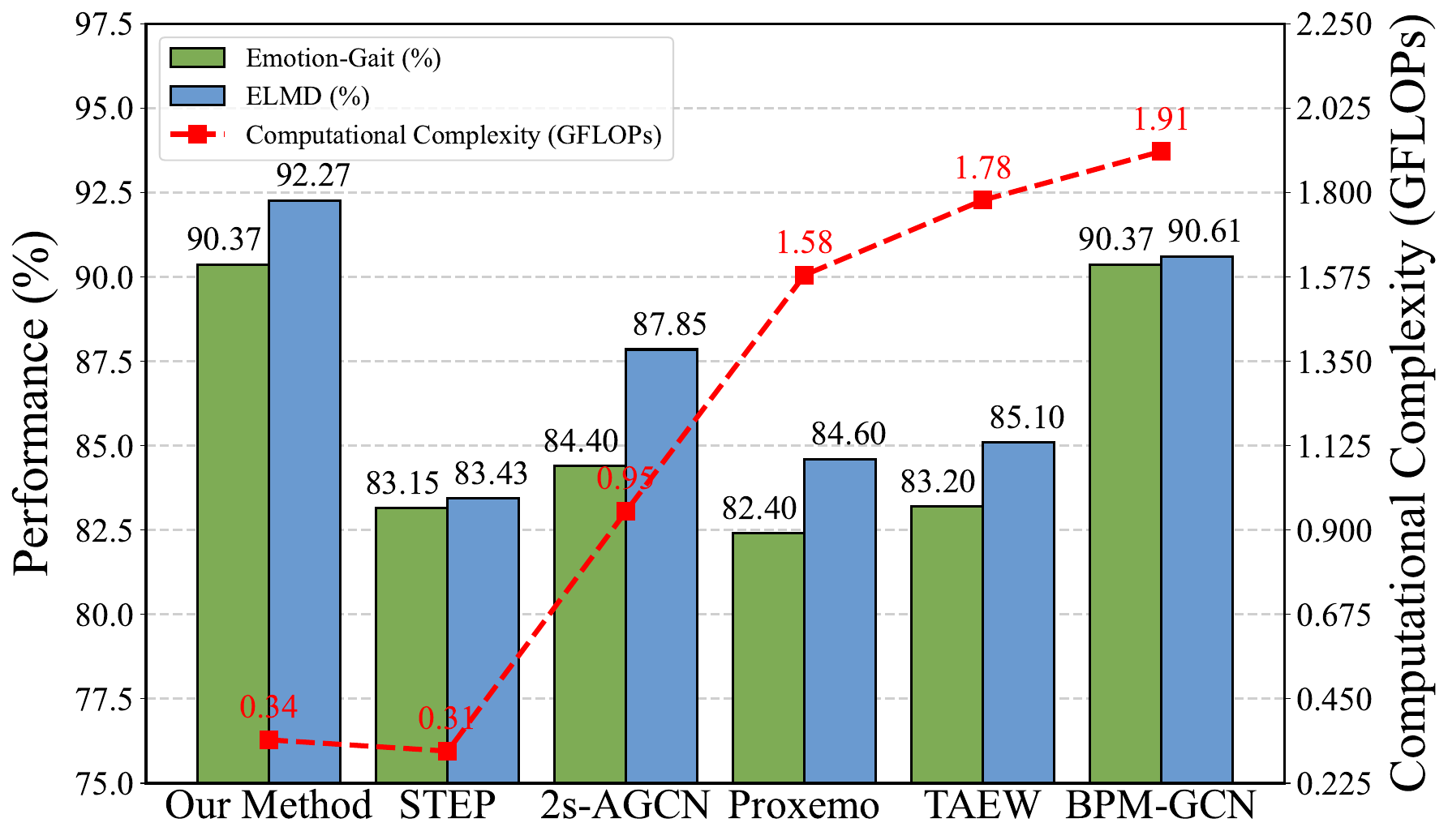}
    \caption{\textbf{Performance \vs Computational Complexity.} Our method achieves state-of-the-art performance in gait emotion recognition while maintaining computational efficiency during testing.}
    \label{fig: motivation}
\end{figure}

Recently, graph convolution-based methods~\cite{chen2023sta, yin2024msa, zhang2024tt, zhuang2020g} have achieved significant progress, as the skeleton naturally forms a graph structure in non-Euclidean geometric spaces, where each graph node corresponds directly to a skeletal joint. Prior studies typically adopt network architectures similar to those used in action recognition, such as the Graph Convolutional Network (GCN) for spatial typology modeling and the Temporal Convolutional Network (TCN) for temporal dynamic modeling. However, research~\cite{yin2024msa} highlights a key distinction between action recognition and emotion recognition: While action recognition primarily focuses on local joint movements, emotion recognition involves classifying emotions based on the overall body state. Consequently, TCN alone is insufficient for capturing long-range temporal dependencies, which are crucial for emotion recognition. 

On the other hand, some methods employ dual-stream network architectures to capture more comprehensive emotion features. For example, BPM-GCN~\cite{zhai2024looking} is the first method that fuses human posture information and higher-order motion features by dynamic weight addition. However, this fusion manner fails to fully capture the correlation between different streams and does not effectively leverage their complementary information for improved recognition. Therefore, we employ the Transformer to align and complement features across different streams based on semantic similarity, forming a more comprehensive emotion representation.

Transformer~\cite{kim2021bert, vaswani2017attention} has become the dominant model in natural language processing (NLP) due to its powerful long-range modeling capabilities. Inspired by the success of NLP, numerous studies~\cite{carion2020end, chen2021crossvit, dosovitskiy2020image, liu2021swin, zamir2022restormer, zhou2023fourmer} have extended its application to computer vision, achieving state-of-the-art performance. Building on this progress, researchers have explored its potential in video understanding tasks, including action recognition~\cite{lee2023cast, wu2024macdiff}, video generation~\cite{guo2024i2v, jiang2024videobooth}, and spatiotemporal grounding~\cite {wang2023efficient, wasim2024videogrounding}. Recently, large multimodal models~\cite{li2024monkey, liu2024improved} have also adopted transformer-based architectures. Motivated by these advancements, we incorporate transformers into skeleton-based gait emotion recognition to capture global temporal representations and bridge information gaps between different streams (\ie, posture and motion).

Specifically, in this paper, we propose a novel framework, called \textbf{CGTGait}, which is the first work to integrate GCNs and Transformers for skeleton-based gait emotion recognition. CGTGait adopts a dual-stream architecture comprising posture and motion streams, where each stream utilizes stacked CGT blocks to capture both frame-level spatial topology and global temporal information. Additionally, we incorporate a contrastive learning strategy within each CGT block to enhance feature discriminability and robustness by constraining the distance between confident and ambiguous samples. To further bridge spatiotemporal information gaps between the two streams, we introduce a Bidirectional Cross-Stream Fusion (BCSF) module, which utilizes the temporal cross-transformer and dynamic spatial attention to facilitate the exchange of complementary information between posture and motion features. The former facilitates long-range temporal dependencies and context awareness across frames between different streams, while the latter adaptively emphasizes key spatial typology information, ensuring a more comprehensive understanding of emotional expressions. We conduct extensive experiments on two public datasets (Emotion-Gait~\cite{bhattacharya2020step} and ELMD~\cite{bhattacharya2020take}) to validate the effectiveness of CGTGait. The results demonstrate that our method achieves superior or at least competitive performance with state-of-the-art methods while significantly reducing computational complexity by approximately 82.2\% (\ie, only requiring 0.34G FLOPs) during testing (as shown in \autoref{fig: motivation}).

The main contributions of this paper can be summarized as follows:
\begin{itemize}
    \item We propose CGTGait, a novel gait emotion recognition method that couples graphs and transformers to capture more discriminative spatiotemporal representations. To the best of our knowledge, CGTGait is the first method to leverage the global temporal modeling capability of transformers for skeleton-based gait emotion recognition.
    \item We introduce BCSF, a Bidirectional Cross-Stream Fusion module that dynamically aligns posture and motion representations using the temporal cross-transformer and dynamic spatial attention, effectively bridging spatiotemporal gaps and facilitating the exchange of complementary information between posture and motion features.
    \item Our method achieves state-of-the-art or comparable performance on two datasets (Emotion-Gait~\cite{bhattacharya2020step} and ELMD~\cite{bhattacharya2020take}), while significantly reducing test computational complexity by approximately 82.2\%, requiring only 0.34G FLOPs.
\end{itemize}

%% file: sec/2_relat.tex
\section{Related Work}
In this section, we review the most relevant studies on gait emotion recognition and the application of Transformers in skeleton-based action recognition.

\begin{figure*}
    \centering
    \includegraphics[width=0.9\linewidth]{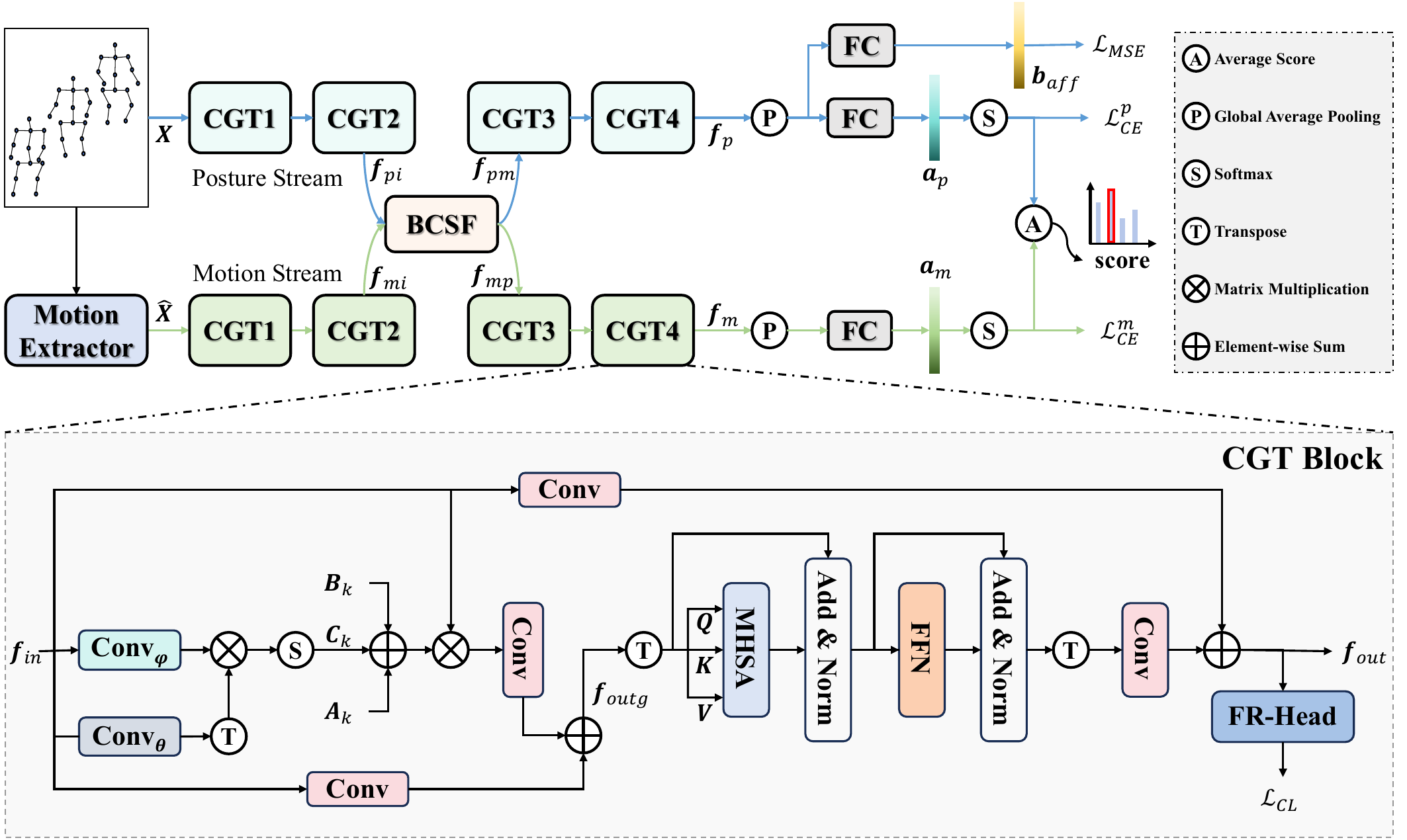}
    \caption{(a) \textbf{Overview of our CGTGait.} CGTGait is a dual-stream architecture, comprising posture and motion streams. Each stream incorporates four CGT blocks to extract global spatiotemporal information. Additionally, the BCSF module bidirectionally aligns posture and motion features using a temporal cross-transformer and dynamic spatial attention. Finally, the test score is obtained by averaging the scores from both streams. (b) \textbf{Detailed structure of the CGT block.} A CGT block consists of a GCN for capturing discriminative spatial topology information and a Transformer for modeling global temporal dependencies.}
    \label{fig: overview}
\end{figure*}

\subsection{Gait Emotion Recognition}
Early studies~\cite{crenn2016body, karg2010recognition, omlor2007extraction, venture2014recognizing, chiu2018emotion} employ traditional machine learning methods with hand-crafted features to extract emotional information, such as Principal Component Analysis (PCA), Linear Discriminant Analysis (LDA), and Support Vector Machines (SVM). For example, Li \etal~\cite{li2016identifying} identify emotions by leveraging the discrete Fourier transform and statistical principles. However, hand-crafted feature modeling is complex, and variations across different datasets significantly hinder its generalization ability.

Recently, the rapid advancement of deep learning has significantly contributed to the development of gait emotion recognition. Based on their encoding methods, deep learning models can be broadly categorized into sequence-based, image-based, and graph-based methods~\cite{chen2023ast, hu2022tntc}. Sequence-based methods~\cite{bhattacharya2020take, randhavane2019identifying, randhavane2019learning} typically employ Long Short-Term Memory (LSTM) or Gated Recurrent Unit (GRU) to model inter-frame information in skeleton sequences. Image-based methods~\cite{narayanan2020proxemo, hu2022tntc} transform skeleton sequences into images and then leverage CNN to extract hierarchical emotional representations. However, both sequence- and image-based methods struggle to capture interactions between distant, non-adjacent joints due to the lack of spatial modeling or the limited receptive field of convolutional kernels. As skeleton structures naturally form graphs in non-Euclidean geometric spaces, the core idea of graph-based methods~\cite{bhattacharya2020step, chen2023ast, lu2023epic, yin2024msa, chen2023sta, lima2024st, zhu2024temporal, zhai2024looking, zhang2024tt, zhuang2020g,zeng2025gaitcycformer} is primarily to leverage GCN to model spatial topologies and TCN to capture temporal motion dynamics. For example, Chen and Sun~\cite{chen2023sta} present STA-GCN, which adaptively aggregates implicit and explicit spatial features along with multi-scale joint motion information. Additionally, Zhai \etal~\cite{zhai2024looking} propose a dual-steam network BPM-GCN to extract emotion features from both posture and motion streams. However, existing studies fail to fully exploit the complementary information between different streams, leading to the loss of discriminative features.

\subsection{Transformers in Skeleton-based Action Recognition}
The Transformer model excels at capturing global temporal information, in contrast to TCN, which has a limited receptive field. Several works~\cite{zhang2021stst, do2024skateformer, wang20233mformer, qu2024llms, plizzari2021spatial} have leveraged transformers for skeleton-based action recognition to extract more comprehensive spatiotemporal features. For example, Zhou \etal~\cite{zhou2022hypergraph} introduce Hyperformer, which integrates skeleton structure information into the Transformer so as to capture inherent high-order joint relationships. Similarly, Wu \etal~\cite{wu2024frequency} propose FreqMixFormer, which enhances spatiotemporal correlation extraction by integrating frequency features with spatial representations. Additionally, other works~\cite{chi2024infogcn++, geng2024hierarchical,guo2024mg,zhu2022mlst,pang2023self} have explored hybrid architectures that combine GCNs and Transformers for improved spatiotemporal feature extraction. Notably, Chi \etal~\cite{chi2024infogcn++} integrate NeuralODEs with Transformer to enable online action recognition. 

In this paper, we propose CGTGait, the first framework to collaboratively integrate both GCNs and Transformers for extracting more discriminative features in skeleton-based gait emotion recognition. Furthermore, we employ the cross-transformer to enhance cross-frame context understanding between different streams.

%% file: sec/3_method.tex
\section{Method}
In this section, we first present the overall architecture of our method in \autoref{sec: overview}. Next, we describe the detailed structure of the collaborative graph and transformer in \autoref{sec: CGT}. Finally, we provide a comprehensive explanation of the bidirectional cross-stream fusion module in \autoref{sec: BCSF}. 

\subsection{Overview}
\label{sec: overview}
We represent a skeleton sequence with $\mathtt{T}$ frames as $\boldsymbol{X} \in \mathbb{R}^{\mathtt{C}_{in}\times \mathtt{T}\times \mathtt{N}}$, where $\mathtt{C}_{in}$ and $\mathtt{N}$ denote the dimensionality of each joint and the number of joints per frame, respectively. As shown in \autoref{fig: overview}, our proposed CGTGait consists of two streams: the posture stream and the motion stream. The posture stream directly takes the original sequence $\boldsymbol{X}$ as input and employs four stacked Collaborative Graph and Transformer (CGT) blocks to extract the human posture spatiotemporal feature $\boldsymbol{f}_{p} \in \mathbb{R} ^ {\mathtt{C}_{p}\times \mathtt{T}_{p} \times \mathtt{N}}$ (\eg, joint angles and distances). The motion stream first applies a motion extractor $\mathcal{M}(\cdot)$ to generate the motion sequence $\boldsymbol{\hat{X}} \in \mathbb{R}^{\mathtt{C}'_{in}\times \mathtt{T}\times \mathtt{N}}$ (\ie, $\mathtt{C}'_{in} = 8$, including three velocity components, overall velocity, three acceleration components, and overall acceleration for each joint). Subsequently, another set of four stacked CGT blocks is employed to capture the high-order motion feature $\boldsymbol{f}_{m} \in \mathbb{R} ^ {\mathtt{C}_{m}\times \mathtt{T}_{m} \times \mathtt{N}}$. Additionally, a Bidirectional Cross-Stream Fusion (BCSF) module is introduced after two CGT blocks to align posture and motion features, thereby facilitating the exchange of complementary information between the two streams. Finally, the test score is obtained by averaging the scores from both streams. Furthermore, the posture stream is enhanced with 31 affective features~\cite{zhai2024looking} (\ie, 14 angle features, 9 distance features, and 8 area features) to more effectively capture emotional cues from the gait sequence. Additional details are provided in the following sections.

\subsection{Collaborative Graph and Transformer}
\label{sec: CGT}
\textbf{Motivation.} As discussed in \autoref{sec: intro}, long-range dependencies between gait frames encode distinctive biological information, which is essential for accurate emotion recognition. However, existing methods primarily rely on TCN to capture local temporal details, neglecting long-range temporal dependencies. To address this limitation, we propose the Collaborative Graph and Transformer (CGT) block for comprehensive global spatiotemporal feature extraction.

\textbf{Operation.} The details of the CGT block are illustrated in \autoref{fig: overview}(b). By leveraging the interaction between GCN and Transformer, the CGT block effectively captures frame-level spatial topology and global temporal dependencies, thereby obtaining a more comprehensive and discriminative spatiotemporal representation. Specifically, our CGT block can be formulated as follows: 
\begin{equation}
    \boldsymbol{f}_{out} = \mathcal{T}(\mathcal{G}(\boldsymbol{f}_{in})) + \text{Conv}(\boldsymbol{f}_{in}),
    \label{eq: CGT}
\end{equation}
where $\mathcal{G}(\cdot)$ and $\mathcal{T}(\cdot)$ denote graph modeling and transformer modeling, respectively, and $\text{Conv}(\cdot)$ represents the residual connection. Finally, we incorporate an FR Head~\cite{zhou2023learning} and a contrastive loss $\mathcal{L}_{CL}$ to enhance the discriminative capability and robustness of the output feature $\boldsymbol{f}_{out}$ by minimizing the distance between confident and ambiguous samples.

\textbf{Graph Modeling:} Given an input feature $\boldsymbol{f}_{in} \in \mathbb{R}^{\mathtt{C}_{i} \times \mathtt{T}_{i} \times \mathtt{N}}$, we first compute the self-attention adjacency matrix $\boldsymbol{C}_{k} \in \mathbb{R}^{\mathtt{N}\times \mathtt{N}}$ to establish global joint connections for each sequence, formulated as:
\begin{equation}
    \boldsymbol{C}_{k} = \text{Softmax}\left(\text{Conv}_{\theta}^\top(\boldsymbol{f}_{in}) \otimes \text{Conv}_{\varphi}(\boldsymbol{f}_{in})\right),
\end{equation}
where $\text{Conv}_{\theta}(\cdot)$ and $\text{Conv}_{\varphi}(\cdot)$ denote two CNNs with a kernel size of $1 \times 1$. $\text{Softmax}(\cdot)$ and the symbol of $\otimes$ represent the softmax function and matrix multiplication, respectively. Subsequently, the output feature $\boldsymbol{f}_{outg} \in \mathbb{R}^{\mathtt{C}_o \times \mathtt{T}_i \times \mathtt{N}}$ of graph modeling process is formulated as follows:
\begin{equation}
    \begin{aligned}
    \boldsymbol{f}_{outg} = 
    & \sum_{k=1}^{\mathtt{K}_v}\text{Conv}_k\left(\boldsymbol{f}_{in} \otimes (\boldsymbol{A}_{k} + \boldsymbol{B}_{k} + \boldsymbol{C}_{k})\right) \\
    & + \text{Conv}(\boldsymbol{f}_{in})
    \end{aligned}.
\end{equation}
Here, $\boldsymbol{A}_{k} \in \mathbb{R}^{\mathtt{N}\times \mathtt{N}}$ is a predefined adjacency matrix that represents the human physical structure based on prior knowledge, while $\boldsymbol{B}_{k} \in \mathbb{R}^{\mathtt{N}\times \mathtt{N}}$ is a learnable adjacency matrix, initialized as zero and updated in an end-to-end manner. $\mathtt{K}_v$ denotes the number of graph subsets. Through graph modeling, we obtain frame-level global spatial topology information.

\textbf{Transformer Modeling:} The transformer modeling is designed to capture global relationships among different frames, consisting of a multi-head self-attention (MHSA) mechanism, a feedforward network (FFN), and a temporal downsample CNN. Specifically, the feature $\boldsymbol{f}_{outg}$ is first linearly transformed into the query $\boldsymbol{Q} \in \mathbb{R}^{\mathtt{N} \times \mathtt{T}_i \times \mathtt{C}_t}$, key $\boldsymbol{K} \in \mathbb{R}^{\mathtt{N} \times \mathtt{T}_i \times \mathtt{C}_t}$, and value $\boldsymbol{V} \in \mathbb{R}^{\mathtt{N} \times \mathtt{T}_i \times \mathtt{C}_t}$ as follows:
\begin{equation}
    \left[\boldsymbol{Q}, \boldsymbol{K}, \boldsymbol{V}\right] = \boldsymbol{f}_{outg}^\top \otimes \left[W_{Q}, W_{K}, W_{V}\right],
    \label{eq: qkv}
\end{equation}
where $W_Q$, $W_K$, and $W_V \in \mathbb{R}^{\mathtt{C}_o \times \mathtt{C}_t} $ are learnable weight matrices. The MHSA mechanism then splits the $\boldsymbol{Q}$, $\boldsymbol{K}$, and $\boldsymbol{V}$ into $\mathtt{h}$ heads and performs self-attention in parallel across these heads. The output from all heads is subsequently concatenated and linearly projected to form the final output. Consequently, the complete transformer modeling is formulated as:
\begin{equation}
    \begin{aligned}
        &\text{MHSA} \left(\boldsymbol{f}_{outg}^\top \right) = \text{Softmax}\left(\frac{\boldsymbol{Q} \otimes \boldsymbol{K}^\top}{\sqrt{\mathtt{C}_t/ \mathtt{h}}}\right) \otimes \boldsymbol{V} \\
        &\boldsymbol{f}_{att} = \text{LN}\left(\text{MHSA}\left(\boldsymbol{f}_{outg}^\top \right) + \boldsymbol{f}_{outg}^\top \right) \\
        & \boldsymbol{f}_{trans} = \text{LN}\left(\text{FFN}\left(\boldsymbol{f}_{att} \right) + \boldsymbol{f}_{att} \right)
    \end{aligned}.
    \label{eq: h1}
\end{equation}
Here, $\mathtt{h}$ is the number of heads in MHSA, and $\text{LN}(\cdot)$ denotes the LayerNorm function. Finally, we apply a point-wise CNN to map $\boldsymbol{f}_{trans}$. To reduce information redundancy and computational complexity, the stride of the point-wise convolution in the last two CGT blocks is set to $2 \times 1$. Through transformer modeling, we extract global temporal dependencies.

\begin{figure}
    \centering
    \includegraphics[width=0.8\linewidth]{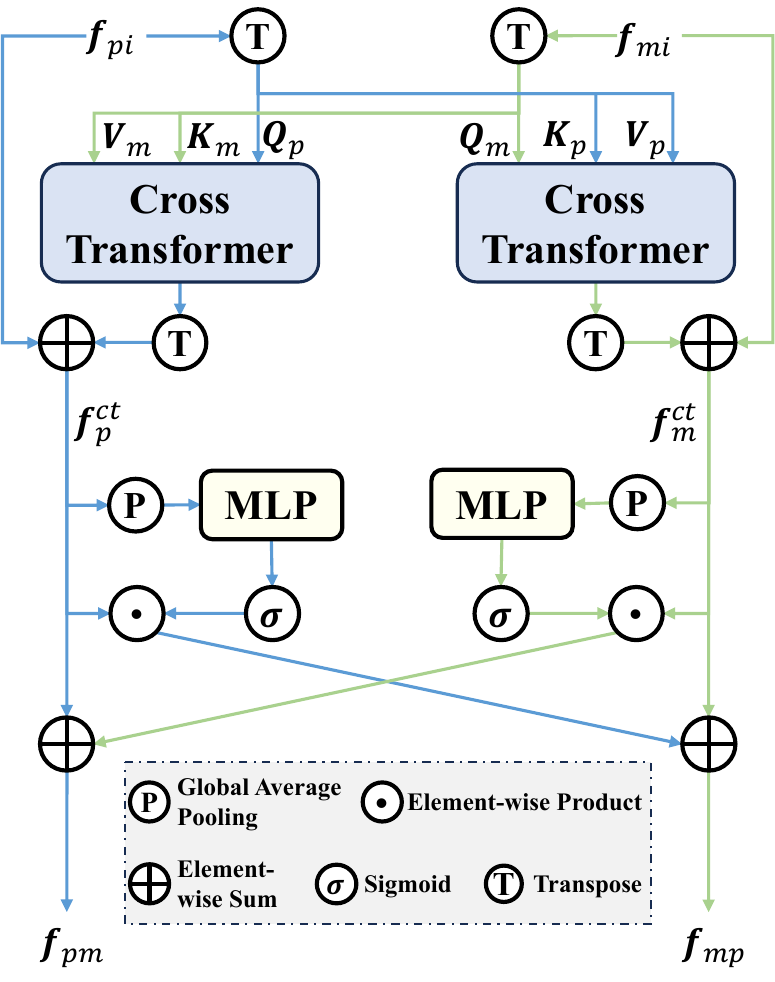}
    \caption{\textbf{The network structure of our proposed BCSF module.} BCSF consists of two main components: the temporal cross-transformer and dynamic spatial attention. The temporal cross-transformer promotes context awareness across frames, while the dynamic spatial attention highlights key spatial topology information.}
    \label{fig: BCSF}
\end{figure}

\subsection{Bidirectional Cross-Stream Fusion}
\label{sec: BCSF}
\textbf{Motivation.} As a video-based recognition task, existing studies fail to fully utilize the temporal information across different streams, leading to a loss of discriminative features. Motivated by the success of Transformers in multimodal learning~\cite{peebles2023scalable},  we present the Bidirectional Cross-Stream Fusion (BCSF) module to enhance information exchange between two streams, with a particular focus on context awareness across frames for more comprehensive emotion recognition. 

\textbf{Operation.} 
As shown in \autoref{fig: BCSF}, the BCSF module comprises two temporal cross-transformers and two dynamic spatial attentions, effectively bridging spatiotemporal gaps and facilitating cross-stream information exchange. The BCSF process can be formulated as:
\begin{equation}
    \begin{aligned}
        \boldsymbol{f}_{pm} &= \mathcal{TSF}_p(\boldsymbol{f}_{pi}, \boldsymbol{f}_{mi}) \\
        \boldsymbol{f}_{mp} &= \mathcal{TSF}_m(\boldsymbol{f}_{mi}, \boldsymbol{f}_{pi})
    \end{aligned},
\end{equation}
where $\mathcal{TSF}_p$ and $\mathcal{TSF}_m$ denote the spatiotemporal fusion of the posture and motion streams, respectively. This design allows the network to attend to human posture while effectively capturing motion dynamics, thereby enhancing emotion recognition performance.

Given the symmetry of dual-stream fusion, we introduce the detailed structure using $\mathcal{TSF}_p$ as an example.

\textbf{Temporal Fusion:} Similar to \autoref{eq: qkv}, we first linearly project the two input features $\boldsymbol{f}_{pi}$ and $\boldsymbol{f}_{mi} \in \mathbb{R}^{\mathtt{C}_i' \times \mathtt{T}_i' \times \mathtt{N}}$ into three matrices: the posture stream's query $\boldsymbol{Q}_p$, the motion stream's key $\boldsymbol{K}_m$, and the motion stream's value $\boldsymbol{V}_m$. A multi-head cross-attention (MHCA) is then applied to facilitate dual-stream feature learning and fusion, as follows:
\begin{equation}
    \boldsymbol{CrossAttn}_{pm} = \text{Softmax}\left(\frac{\boldsymbol{Q}_p \otimes \boldsymbol{K}_m^\top}{\sqrt{\mathtt{C}_k / \mathtt{h}}}\right) \otimes \boldsymbol{V}_m,
    \label{eq: tf1}
\end{equation}
where $\mathtt{C}_k$ is the channel number of $\boldsymbol{K}_m$. Next, a feedforward network (FFN) is employed to further integrate the extracted cross-stream features. To enhance learning while preserving the original stream features, residual connections are incorporated. This process can be formulated as follows:
\begin{equation}
    \begin{aligned}
        \boldsymbol{f}_p' & = \text{LN}\left(\boldsymbol{CrossAttn}_{pm} + \boldsymbol{f}_{pi}^\top\right) \\
        \boldsymbol{f}_p^{ct} &= \left(\text{LN}\left(\text{FFN}\left(\boldsymbol{f}_p'\right) + \boldsymbol{f}_p' \right)\right)^\top + \boldsymbol{f}_{pi}
    \end{aligned}.
    \label{eq: tf2}
\end{equation}
Here, $\boldsymbol{f}_p^{ct} \in \mathbb{R}^{\mathtt{C}_i' \times \mathtt{T}_i' \times \mathtt{N}}$ is the output of temporal fusion. Similarly, applying \autoref{eq: tf1} and \autoref{eq: tf2} to the motion stream yields its temporal fusion feature $\boldsymbol{f}_m^{ct} \in \mathbb{R}^{\mathtt{C}_i' \times \mathtt{T}_i' \times \mathtt{N}}$. This temporal fusion strategy effectively aligns the temporal features of the posture and motion streams, enabling the extraction of complementary cross-stream features.

\textbf{Spatial Fusion:} First, the feature $\boldsymbol{f}_m^{ct}$ undergoes global average pooling (GAP) along the channel and temporal dimensions to extract the spatial joint topology feature across the entire sequence. Then, a multilayer perceptron (MLP) with a sigmoid activation function is applied to generate the spatial joint attention $\boldsymbol{SAttn}_m \in \mathbb{R}^{N}$, which is formulated as:
\begin{equation}
    \boldsymbol{SAttn}_m = \sigma(\text{MLP}(\text{P}(\boldsymbol{f}_m^{ct}))),
    \label{eq: sf1}
\end{equation}
where $\sigma(\cdot)$ and $\text{P}(\cdot)$ denote the sigmoid function and GAP operation, respectively. The values in $\boldsymbol{SAttn}_m$ represent the relative importance of joints in the motion stream. Finally, the weighted feature based on $\boldsymbol{SAttn}_m$ and $\boldsymbol{f}_m^{ct}$ is added to $\boldsymbol{f}_p^{ct}$ to produce the final spatial fusion feature $\boldsymbol{f}_{pm}  \in \mathbb{R}^{\mathtt{C}_i' \times \mathtt{T}_i' \times \mathtt{N}}$, which is formulated as:
\begin{equation}
    \boldsymbol{f}_{pm} = \boldsymbol{f}_m^{ct} \cdot \boldsymbol{SAttn}_m + \boldsymbol{f}_p^{ct}.
    \label{eq: sf2}
\end{equation}
Similarly, by applying \autoref{eq: sf1} and \autoref{eq: sf2}, we obtain the another output $\boldsymbol{f}_{mp} \in \mathbb{R}^{\mathtt{C}_i' \times \mathtt{T}_i' \times \mathtt{N}}$. This spatial fusion manner focuses on the key spatial joint information for emotion recognition, promoting the interaction between dual-stream spatial features.
 
\subsection{Training Objective}
As shown in \autoref{fig: overview}, three types of loss functions are employed to train our network CGTGait: classification loss, affective distillation loss, and FR contrastive loss.

\textbf{Classification Loss:} For both the posture and motion streams, we use cross-entropy loss for classification supervision, formulated as:
\begin{equation}
    \mathcal{L}_{CE} = \mathcal{L}_{CE}^p (\boldsymbol{y}, \boldsymbol{p}^p) + \mathcal{L}_{CE}^m (\boldsymbol{y}, \boldsymbol{p}^m),
\end{equation}
where $\boldsymbol{y}$ is the ground truth. $\boldsymbol{p}^{p}$ and $\boldsymbol{p}^{m}$ denote the probability scores of the posture and motion streams, respectively.

\textbf{Affective Distillation Loss:} As mentioned in \autoref{sec: overview}, to enrich the posture feature with affective information and improve discriminative capability, we apply the affective distillation loss, formulated as:
\begin{equation}
    \mathcal{L}_{MSE} = \Vert \boldsymbol{b}_{aff} - \boldsymbol{y}_a \Vert^2_2,
\end{equation}
where $\boldsymbol{y}_a$ is the ground truth for the affective feature $\boldsymbol{b}_{aff}$.

\textbf{FR Contrastive Loss:} As mentioned in \autoref{sec: CGT}, we apply contrastive loss within each CGT block to refine the feature. The total FR contrastive loss is formulated as: 
\begin{equation}
    \mathcal{L}_{FR} = \sum_{i=1}^4 \lambda_i \cdot (\mathcal{L}_{CL}^{pi} + \mathcal{L}_{CL}^{mi}),
    \label{eq: lambda}
\end{equation}
where $\mathcal{L}_{CL}^{pi}$ and $\mathcal{L}_{CL}^{mi}$ denote the contrastive loss in the $i$-th CGT block of the posture and motion streams, respectively. $\lambda_i$ is the balancing coefficient.

Thus, the overall learning objective can be formulated as:
\begin{equation}
    \mathcal{L} = \mathcal{L}_{CE} + \mathcal{L}_{MSE} + \mathcal{L}_{FR}.
\end{equation}

%% file: sec/4_exper.tex
\section{Experiments}
In this section, we first introduce the used datasets in \autoref{sec: datasets}. Then, we describe the implementation details in \autoref{sec: implementation}. Next, we evaluate the effectiveness of our proposed CGTGait through performance experiments conducted on two datasets, presented in \autoref{sec: quantitative} and \autoref{sec: qualitative}, respectively.  Finally, we perform ablation studies to verify the positive impact of each component in CGTGait, detailed in \autoref{sec: ablation}.

\subsection{Datasets}
\label{sec: datasets}
\textbf{Emotion-Gait.} The Emotion-Gait~\cite{bhattacharya2020step} is a widely used dataset containing 2,177 gait samples across four emotion classes: happy, sad, angry, and neutral. Of these, 1,835 samples, each consisting of 240 frames, are sourced from the Edinburgh Locomotion MOCAP Database~\cite{Habibie2017ARV}. The remaining 342 samples, with varying frame lengths (ranging from 27 to 75), are collected by Bhattacharya \etal~\cite{bhattacharya2020step}. In all our experiments, we divide the Emotion-Gait dataset into training and test sets using a 9:1 ratio.

\textbf{ELMD.} The ELMD~\cite{bhattacharya2020take} consists of 1,835 gait sequences, each containing 240 frames, with four emotion labels (\ie, happy, sad, angry, and neutral) annotated by multiple participants. As described above, ELMD is divided into training and test sets following a 9:1 ratio.

\begin{table}[!t]
    \begin{center}
        \resizebox{0.90\linewidth}{!}{
        \begin{tabular}{c|ccc}
            \toprule
            \textbf{Datasest} & \textbf{Method} & \textbf{Publication} & \textbf{Accuracy} \\ \midrule
            \multirow{18}{*}{\makecell[c]{ Emotion \\ -Gait}} & LSTM~\cite{randhavane2019learning} & ISMAR'19 & 75.10 \\
             & 2s-AGCN~\cite{shi2019two} & CVPR'19 & 84.40 \\
             & G-GCSN~\cite{zhuang2020g} & ACCV'20 & 81.50 \\
             & Proxemo~\cite{narayanan2020proxemo} & IROS'20 & 82.40 \\
             & STEP~\cite{bhattacharya2020step} & AAAI'20 & 83.15 \\
             & TAEW~\cite{bhattacharya2020take} & ECCV'20 & 83.20 \\
             & TNTC~\cite{hu2022tntc} & ICASSP'22 & 85.97 \\
             & STA-GCN~\cite{chen2023sta} & ICME'23 & 86.20 \\
             & CAGE~\cite{lu2023see} & AAAI'23 & 79.59 \\
             & TT-GCN~\cite{zhang2024tt} & TCSS'24 & 80.11 \\
             & AP-Gait~\cite{yumeng2024affective} & MTA'24 & 85.20 \\
             & AST-GCN~\cite{chen2023ast} & TCSVT'24 & 88.04 \\
             & ST-Gait++~\cite{lima2024st} & CVPRW'24 & 87.50 \\
             & EPIC~\cite{lu2023epic} & JBHI'24 & 89.66 \\
             & MAHANN~\cite{zhang2024hierarchical} & Physic A'24 & \underline{90.20} \\
             & BPM-GCN~\cite{zhai2024looking} & TAFFC'24 & \textbf{90.37} \\
             & MS-GCN~\cite{zou2025occluded} & CMC'25 & 90.00 \\
             & TS-ST~\cite{li2025gait} & Sensors'25 & 84.15 \\
             & GaitCycFormer~\cite{zeng2025gaitcycformer} & AAAI'25 & 86.30 \\ \cmidrule{2-4}
             & \textbf{Ours} & IJCB'25 & \textbf{90.37} \\ \midrule
            \multirow{8}{*}{ELMD} & LSTM~\cite{randhavane2019learning} & ISMAR'19 & 78.50 \\
             & 2s-AGCN~\cite{shi2019two} & CVPR'19 & 87.85 \\
             & Proxemo~\cite{narayanan2020proxemo} & IROS'20 & 84.60 \\
             & STEP~\cite{bhattacharya2020step} & AAAI'20 & 83.43 \\
             & TAEW~\cite{bhattacharya2020take} & ECCV'20 & 85.10 \\
             & BPM-GCN~\cite{zhai2024looking} & TAFFC'24 & \underline{90.61} \\
             & MS-GCN~\cite{zou2025occluded} & CMC'25 & 89.60 \\ \cmidrule{2-4}
             & \textbf{Ours} & IJCB'25 & \textbf{92.27} \\
            \bottomrule
        \end{tabular}}
    \end{center}
    \caption{Performance comparisons on Emotion-Gait~\cite{bhattacharya2020step} and ELMD~\cite{bhattacharya2020take} datasets. The best and second-best results are highlighted in \textbf{bold} and \underline{underlined}, respectively.}
    \label{tab: Emotion-Gait + ELMD}
\end{table}

\begin{table}[!t]
    \begin{center}
    \resizebox{0.85\linewidth}{!}{
    	\begin{tabular}{cccc}
    		\toprule
    		\textbf{Method} & \textbf{Paramter} & \textbf{FLOPs} & \textbf{ELMD} \\
    		\midrule
            STEP~\cite{bhattacharya2020step} & 0.71M & 0.31G & 83.43 \\
            2s-AGCN~\cite{shi2019two} & 3.44M & 0.95G & 87.85 \\
            Proxemo~\cite{narayanan2020proxemo} & 0.08M & 1.58G & 84.60 \\
            TAEW~\cite{bhattacharya2020take} & 40.43M & 1.78G & 85.10 \\
            BPM-GCN~\cite{zhai2024looking} & 7.27M & 1.91G & 90.61 \\
            CGTGait & 2.66M & 0.34G & 92.27 \\
    		\bottomrule
    	\end{tabular}
        }
    \end{center}
	\caption{Comparison of parameter and computational complexity with other methods.}
	\label{tab: param_computation}
\end{table}

\begin{figure*}[!t]
    \centering
    \includegraphics[width=\linewidth]{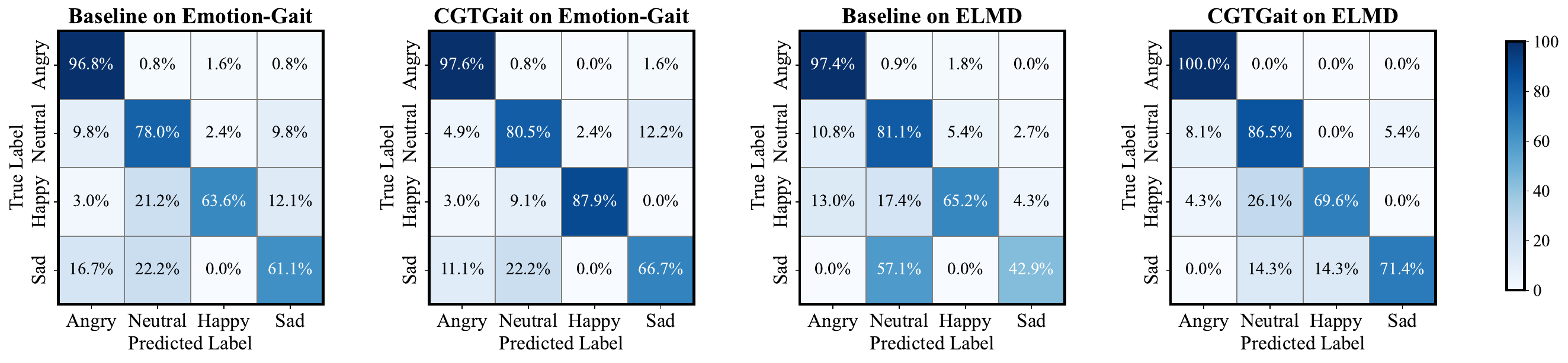}
    \caption{\textbf{Visualization of confusion matrices comparing the baseline model and CGTGait on the Emotion-Gait and ELMD datasets.}}
    \label{fig: confusion}
\end{figure*}

\subsection{Implementation Details} 
\label{sec: implementation}
\textbf{Hyper-parameters.} The input skeleton sequence size $\mathtt{C}_{in} \times \mathtt{T} \times \mathtt{N}$ is uniformly set to $3 \times 48 \times 16$ on two datasets. Following the protocol in~\cite{shi2019two}, we apply random rotation and translation as data augmentation strategies during training. The input-output channels for the four CGT blocks in both the posture and motion streams are set to $(3, 64), (64, 64), (64, 128)$ and $(128, 256)$, respectively. The number of heads, $\mathtt{h}$ in \autoref{eq: h1} and \autoref{eq: tf1}, is set to 8. The balancing coefficients $[\lambda_1, \lambda_2, \lambda_3, \lambda_4]$ in \autoref{eq: lambda} are defined as $[0.1,0.2,0.5,1.0]$.

\textbf{Training details.} All experiments are conducted on an NVIDIA GeForce RTX 3090 GPU using the Pytorch~\cite{paszke2019pytorch} framework. SGD with a momentum of 0.9 is employed as the optimizer. The initial learning rate is set to 0.01 and decays by a factor of 0.1 every 30 epochs. Our method is trained for 80 epochs with a batch size of 32.

\subsection{Performance Comparison}
\label{sec: quantitative}
As shown in \autoref{tab: Emotion-Gait + ELMD} and \autoref{tab: param_computation}, we compare the performance and testing computational complexity of our proposed CGTGait with existing methods on the Emotion-Gait~\cite{bhattacharya2020step} and ELMD~\cite{bhattacharya2020take} datasets. Our comparison leads to the following key findings: 1) CGTGait achieves state-of-the-art or highly competitive performance compared to existing methods across two datasets, demonstrating its effectiveness. 2) Compared to STEP~\cite{bhattacharya2020step} with the minimum computational complexity, our method shows a performance improvement of 7.22\% and 8.84\% on Emotion-Gait and ELMD, respectively, with a minimal increase in computational complexity of only 0.03G FLOPs (0.34G \vs 0.31G), which is acceptable. 3) When compared to the previous best method, BMP-GCN~\cite{zhai2024looking}, CGTGait demonstrates a significant reduction in computational complexity. Specifically, it requires only 0.34G FLOPs (approximately an 82.2\% decrease), while improving accuracy by 1.66\% on the ELMD dataset, highlighting its superior efficiency and performance.

\subsection{Visualization}
\label{sec: qualitative}
In \autoref{fig: confusion}, we present confusion matrices comparing our method with the baseline model, which removes BCSF from CGTGait and replaces the Transformer with a standard TCN (kernel size $9 \times 1$). The visualizations indicate that CGTGait significantly improves the recognition accuracy across all emotional categories on both datasets. Furthermore, the macro F1 score increases substantially by 0.080 (from 0.766 to 0.846) and 0.095 (from 0.740 to 0.835), respectively. Notably, on Emotion-Gait, CGTGait outperforms the baseline by 24.3\% (87.9\% \vs 63.6\%) in the happy category. Similarly, on ELMD,  it achieves a 28.5\% (71.4\% \vs 42.9\%) improvement in recognizing sad emotions. These results highlight the effectiveness of our method in addressing the challenge of uneven data distribution (\eg, only 15.3\% of happy emotions and 9.1\% of sad emotions on Emotion-Gait).

\begin{table}[!t]
    \begin{center}
    	\begin{tabular}{ccccc}
    		\toprule
    		\textbf{Baseline} & \textbf{CGT} & \textbf{BCSF} & \textbf{Emotion-Gait}  & \textbf{ELMD}\\
    		\midrule
    		\checkmark & & & 85.32 & 88.95\\
    		\checkmark & \checkmark & & 86.24 & 90.06 \\
    		\checkmark & & \checkmark & 88.07 & 91.16\\
    		\checkmark & \checkmark & \checkmark & \textbf{90.37} & \textbf{92.27}\\
    		\bottomrule
    	\end{tabular}
    \end{center}
	\caption{The ablation study of CGT and BSCF on Emotion-Gait and ELMD datasets.}
	\label{tab: ablation_component}
\end{table}

\subsection{Ablation Studies}
\label{sec: ablation}
\textbf{Effectiveness of Proposed Components in CGTGait.} As shown in \autoref{tab: ablation_component}, we evaluate the effectiveness of the proposed components in CGTGait. The baseline model is the same as in \autoref{sec: qualitative}. Our findings reveal that: 1) Both CGT and BCSF contribute to performance improvements (+0.88\% and +1.11\% for CGT, and +2.75\% and 2.21\% for BCSF), demonstrating their effectiveness. 2) Compared to CGT, BCSF shows a more substantial improvement (+0.88\% \vs +2.75\%, and +1.11\% \vs +2.21\%), highlighting the greater significance of feature interaction between the two streams in enhancing emotion recognition performance. 3) The combination of CGT and BCSF provides a more comprehensive spatiotemporal feature representation, leading to the best overall performance and underscoring their complementarity.

\begin{table}[!t]
    \begin{center}
            \begin{tabular}{ccc}
                \toprule
                \textbf{Strategy} & \textbf{Emotion-Gait} & \textbf{ELMD}\\
                \midrule
                GCN-Transformer & \textbf{90.37} & \textbf{92.27}\\
                Transformer-GCN  & 87.61 & 91.71\\
                Parallel & 86.70 & 90.61\\
                \bottomrule
            \end{tabular}
    \end{center}
    \caption{The ablation study of spatiotemporal modeling strategy in CGT on Emotion-Gait and ELMD datasets. GCN-Transformer and Transformer-GCN represent two different serial connection orders between GCN and Transformer, respectively.}
    \label{tab: ablation_order_cgt}
\end{table}

\textbf{Impact of Spatiotemporal Modeling Strategy in CGT.} In \autoref{tab: ablation_order_cgt}, we examine the effect of different spatiotemporal modeling strategies in CGT on emotion recognition. The results demonstrate that the serial connection of GCN and Transformer, particularly in the GCN-Transformer configuration, outperforms the parallel strategy. Specifically, performance improves from 86.70\% to 90.37\% on Emotion-Gait and from 90.61\% to 92.27\% on ELMD. These findings suggest that the serial connection is more effective in capturing rich and discriminative features, resulting in better emotion recognition performance.

\begin{table}[!t]
    \begin{center}
    	\begin{tabular}{c c c c c}
    		\toprule
    		\textbf{Baseline} & \textbf{SF} & \textbf{TF} & \textbf{Emotion-Gait} & \textbf{ELMD}\\
    		\midrule
    		\checkmark & & & 86.24 & 90.06\\
    		\checkmark & \checkmark & & 87.16 & 90.61\\
    		\checkmark & & \checkmark & 89.91 & 91.71\\
    		\checkmark & \checkmark & \checkmark & \textbf{90.37} & \textbf{92.27}\\
    		\bottomrule
    	\end{tabular}
    \end{center}
	\caption{The ablation study of spatiotemporal fusion in BCSF on Emotion-Gait and ELMD datasets. SF and TF represent spatial and temporal fusion, respectively. }
	\label{tab: ablation_bcsf}
\end{table}

\textbf{Effectiveness of Spatiotemporal Fusion in BCSF.} As shown in \autoref{tab: ablation_bcsf}, we investigate the effectiveness of spatiotemporal fusion in BCSF. In this case, the baseline model is CGTGait without BCSF. We find that both spatial and temporal fusions generally improve recognition performance, and their combination yields the best results, highlighting their complementary nature. Additionally, temporal fusion proves more effective than spatial fusion (+3.67\% \vs +0.92\%, and +1.65\% \vs +0.55\%), which also verifies the importance of temporal modeling in gait emotion recognition tasks.

\begin{table}[!t]
    \begin{center}
            \begin{tabular}{ccc}
                \toprule
                \textbf{Position} & \textbf{Emotion-Gait} & \textbf{ELMD}\\
                \midrule
                After CGT1 & 88.07  & 91.16 \\
                After CGT2 & \textbf{90.37} & \textbf{92.27}\\
                After CGT3 & 86.70 & 91.16\\
                After CGT4 & 88.99 & 89.50 \\
                \bottomrule
            \end{tabular}
    \end{center}
    \caption{The ablation study of BCSF position on Emotion-Gait and ELMD datasets.}
    \label{tab: ablation_bcsf_position}
\end{table}

\textbf{Impact of BCSF Position.} As shown in \autoref{tab: ablation_bcsf_position}, we evaluate the impact of BCSF at different positions. The results demonstrate that the best performance is achieved when BCSF is applied after the second CGT block. We analyze the reasons as follows: 1) In early stages, the ability to extract semantic information from the different streams is insufficient, meaning that premature interaction cannot effectively integrate more discriminative semantic information. 2) At a later stage, each stream may lose its distinct emotional information, leading to ineffective integration of meaningful information during subsequent interactions.

\begin{table}[!t]
    \begin{center}
            \begin{tabular}{cccc}
                \toprule
                \textbf{Blocks} & \textbf{Emotion-Gait} & \textbf{ELMD} &\textbf{FLOPs}\\
                \midrule
                3 & 87.16 & 90.06 & \textbf{0.30G} \\
                4 & \textbf{90.37} & \textbf{92.27} & 0.34G\\
                5 & 88.99 & 91.71 & 0.41G \\
                6 & 87.16 & 91.16 & 0.57G\\
                \bottomrule
            \end{tabular}
    \end{center}
    \caption{The ablation study of the number of CGT blocks on Emotion-Gait and ELMD datasets.}
    \label{tab: ablation_cgt_number}
\end{table}

\textbf{Impact of the Number of CGT Blocks.} In \autoref{tab: ablation_cgt_number}, we investigate the influences of the number of CGT Blocks. Our findings indicate that an excessive number of CGT blocks can lead to network overfitting and performance degradation. Consequently, we select a stack of four CGT blocks to achieve optimal performance.

\begin{table}[!t]
    \begin{center}
            \begin{tabular}{cccc}
                \toprule
                \textbf{Head} & \textbf{Emotion-Gait} & \textbf{ELMD} & \textbf{FLOPs} \\
                \midrule
                2 & 88.53 & 91.16 & 0.34G \\
                4 & 89.45 & 91.71 & 0.34G \\
                8 & \textbf{90.37} & \textbf{92.27} & \textbf{0.34G} \\
                16 & 87.61 & 90.06 & 0.34G \\
                \bottomrule
            \end{tabular}
    \end{center}
    \caption{The ablation study of the number of heads $\mathtt{h}$ on Emotion-Gait and ELMD datasets.}
    \label{tab: ablation_h_number}
\end{table}

\textbf{Impact of the Number of Heads $\mathtt{h}$.} 
The influence of the number of heads ($\mathtt{h}$ in \autoref{eq: h1} and \autoref{eq: tf1}) is listed in \autoref{tab: ablation_h_number}. We find that gradually increasing $\mathtt{h}$ can effectively improve performance, indicating that more heads enrich the temporal modeling's capacity. However, an excessive number of heads can lead to performance degradation. We set $\mathtt{h}$ to 4 to achieve optimal performance.

%% file: sec/5_conclu.tex
\section{Conclusion}
This paper introduces a novel approach, CGTGait, which leverages collaborative graphs and transformers for gait emotion recognition. Specifically, we propose CGT to effectively capture both spatial topology and global temporal information. In addition, we present BCSF, a module designed to integrate posture and motion stream features via a temporal cross-transformer and dynamic spatial attention. Extensive experiments demonstrate the effectiveness of our method, achieving state-of-the-art or at least competitive performance while significantly reducing computational complexity.

\textbf{Limitation.} In this paper, we integrate graphs with transformers through a straightforward serial connection strategy, without delving into the potential interactions between the two components. Therefore, a more effective integration strategy enables richer spatiotemporal feature learning. We leave the exploration of such strategies for future work.

%% file: main.bbl
\begin{thebibliography}{10}\itemsep=-1pt

\bibitem{bhattacharya2020step}
U.~Bhattacharya, T.~Mittal, R.~Chandra, T.~Randhavane, A.~Bera, and D.~Manocha.
\newblock Step: Spatial temporal graph convolutional networks for emotion perception from gaits.
\newblock In {\em Proceedings of the AAAI conference on artificial intelligence}, volume~34, pages 1342--1350, 2020.

\bibitem{bhattacharya2020take}
U.~Bhattacharya, C.~Roncal, T.~Mittal, R.~Chandra, K.~Kapsaskis, K.~Gray, A.~Bera, and D.~Manocha.
\newblock Take an emotion walk: Perceiving emotions from gaits using hierarchical attention pooling and affective mapping.
\newblock In {\em European conference on computer vision}, pages 145--163. Springer, 2020.

\bibitem{carion2020end}
N.~Carion, F.~Massa, G.~Synnaeve, N.~Usunier, A.~Kirillov, and S.~Zagoruyko.
\newblock End-to-end object detection with transformers.
\newblock In {\em European conference on computer vision}, pages 213--229. Springer, 2020.

\bibitem{chen2023sta}
C.~Chen and X.~Sun.
\newblock Sta-gcn: Spatial temporal adaptive graph convolutional network for gait emotion recognition.
\newblock In {\em 2023 IEEE International Conference on Multimedia and Expo (ICME)}, pages 1385--1390. IEEE, 2023.

\bibitem{chen2023ast}
C.~Chen, X.~Sun, Z.~Tu, and M.~Wang.
\newblock Ast-gcn: Augmented spatial temporal graph convolutional neural network for gait emotion recognition.
\newblock {\em IEEE Transactions on Circuits and Systems for Video Technology}, 34(6):4581--4595, 2023.

\bibitem{chen2021crossvit}
C.-F.~R. Chen, Q.~Fan, and R.~Panda.
\newblock Crossvit: Cross-attention multi-scale vision transformer for image classification.
\newblock In {\em Proceedings of the IEEE/CVF international conference on computer vision}, pages 357--366, 2021.

\bibitem{chi2024infogcn++}
S.~Chi, H.-g. Chi, Q.~Huang, and K.~Ramani.
\newblock Infogcn++: Learning representation by predicting the future for online skeleton-based action recognition.
\newblock {\em IEEE Transactions on Pattern Analysis and Machine Intelligence}, 2024.

\bibitem{chiu2018emotion}
M.~Chiu, J.~Shu, and P.~Hui.
\newblock Emotion recognition through gait on mobile devices.
\newblock In {\em 2018 IEEE International Conference on Pervasive Computing and Communications Workshops (PerCom Workshops)}, pages 800--805. IEEE, 2018.

\bibitem{crenn2016body}
A.~Crenn, R.~A. Khan, A.~Meyer, and S.~Bouakaz.
\newblock Body expression recognition from animated 3d skeleton.
\newblock In {\em 2016 International Conference on 3D Imaging (IC3D)}, pages 1--7. IEEE, 2016.

\bibitem{deligianni2019emotions}
F.~Deligianni, Y.~Guo, and G.-Z. Yang.
\newblock From emotions to mood disorders: A survey on gait analysis methodology.
\newblock {\em IEEE journal of biomedical and health informatics}, 23(6):2302--2316, 2019.

\bibitem{do2024skateformer}
J.~Do and M.~Kim.
\newblock Skateformer: skeletal-temporal transformer for human action recognition.
\newblock In {\em European Conference on Computer Vision}, pages 401--420. Springer, 2024.

\bibitem{dosovitskiy2020image}
A.~Dosovitskiy, L.~Beyer, A.~Kolesnikov, D.~Weissenborn, X.~Zhai, T.~Unterthiner, M.~Dehghani, M.~Minderer, G.~Heigold, S.~Gelly, et~al.
\newblock An image is worth 16x16 words: Transformers for image recognition at scale.
\newblock {\em arXiv preprint arXiv:2010.11929}, 2020.

\bibitem{geng2024hierarchical}
P.~Geng, X.~Lu, W.~Li, and L.~Lyu.
\newblock Hierarchical aggregated graph neural network for skeleton-based action recognition.
\newblock {\em IEEE Transactions on Multimedia}, 2024.

\bibitem{guo2024i2v}
X.~Guo, M.~Zheng, L.~Hou, Y.~Gao, Y.~Deng, P.~Wan, D.~Zhang, Y.~Liu, W.~Hu, Z.~Zha, et~al.
\newblock I2v-adapter: A general image-to-video adapter for diffusion models.
\newblock In {\em ACM SIGGRAPH 2024 Conference Papers}, pages 1--12, 2024.

\bibitem{guo2024mg}
X.~Guo, Q.~Zhu, Y.~Wang, and Y.~Mo.
\newblock Mg-gct: A motion-guided graph convolutional transformer for traffic gesture recognition.
\newblock {\em IEEE Transactions on Intelligent Transportation Systems}, 2024.

\bibitem{Habibie2017ARV}
I.~Habibie, D.~Holden, J.~Schwarz, J.~Yearsley, and T.~Komura.
\newblock A recurrent variational autoencoder for human motion synthesis.
\newblock In {\em British Machine Vision Conference}, 2017.

\bibitem{halovic2018not}
S.~Halovic and C.~Kroos.
\newblock Not all is noticed: Kinematic cues of emotion-specific gait.
\newblock {\em Human movement science}, 57:478--488, 2018.

\bibitem{hu2022tntc}
C.~Hu, W.~Sheng, B.~Dong, and X.~Li.
\newblock Tntc: Two-stream network with transformer-based complementarity for gait-based emotion recognition.
\newblock In {\em ICASSP 2022-2022 IEEE International Conference on Acoustics, Speech and Signal Processing (ICASSP)}, pages 3229--3233. IEEE, 2022.

\bibitem{jiang2024videobooth}
Y.~Jiang, T.~Wu, S.~Yang, C.~Si, D.~Lin, Y.~Qiao, C.~C. Loy, and Z.~Liu.
\newblock Videobooth: Diffusion-based video generation with image prompts.
\newblock In {\em Proceedings of the IEEE/CVF Conference on Computer Vision and Pattern Recognition}, pages 6689--6700, 2024.

\bibitem{karg2010recognition}
M.~Karg, K.~K{\"u}hnlenz, and M.~Buss.
\newblock Recognition of affect based on gait patterns.
\newblock {\em IEEE Transactions on Systems, Man, and Cybernetics, Part B (Cybernetics)}, 40(4):1050--1061, 2010.

\bibitem{khosravi2023crowd}
M.~R. Khosravi, K.~Rezaee, M.~K. Moghimi, S.~Wan, and V.~G. Menon.
\newblock Crowd emotion prediction for human-vehicle interaction through modified transfer learning and fuzzy logic ranking.
\newblock {\em IEEE transactions on intelligent transportation systems}, 24(12):15752--15761, 2023.

\bibitem{kim2021bert}
S.~Kim, A.~Gholami, Z.~Yao, M.~W. Mahoney, and K.~Keutzer.
\newblock I-bert: Integer-only bert quantization.
\newblock In {\em International conference on machine learning}, pages 5506--5518. PMLR, 2021.

\bibitem{lee2023cast}
D.~Lee, J.~Lee, and J.~Choi.
\newblock Cast: cross-attention in space and time for video action recognition.
\newblock {\em Advances in Neural Information Processing Systems}, 36:79399--79425, 2023.

\bibitem{li2016identifying}
B.~Li, C.~Zhu, S.~Li, and T.~Zhu.
\newblock Identifying emotions from non-contact gaits information based on microsoft kinects.
\newblock {\em IEEE Transactions on Affective Computing}, 9(4):585--591, 2016.

\bibitem{li2025gait}
C.~Li, K.~P. Seng, and L.-M. Ang.
\newblock Gait-to-gait emotional human--robot interaction utilizing trajectories-aware and skeleton-graph-aware spatial--temporal transformer.
\newblock {\em Sensors (Basel, Switzerland)}, 25(3):734, 2025.

\bibitem{li2020deep}
S.~Li and W.~Deng.
\newblock Deep facial expression recognition: A survey.
\newblock {\em IEEE transactions on affective computing}, 13(3):1195--1215, 2020.

\bibitem{li2024monkey}
Z.~Li, B.~Yang, Q.~Liu, Z.~Ma, S.~Zhang, J.~Yang, Y.~Sun, Y.~Liu, and X.~Bai.
\newblock Monkey: Image resolution and text label are important things for large multi-modal models.
\newblock In {\em proceedings of the IEEE/CVF conference on computer vision and pattern recognition}, pages 26763--26773, 2024.

\bibitem{li2021static}
Z.~Li, S.~Yu, E.~B.~G. Reyes, C.~Shan, and Y.-r. Li.
\newblock Static and dynamic features analysis from human skeletons for gait recognition.
\newblock In {\em 2021 IEEE International Joint Conference on Biometrics (IJCB)}, pages 1--7. IEEE, 2021.

\bibitem{lima2024st}
M.~L. Lima, W.~de~Lima~Costa, E.~T. Mart{\'\i}nez, and V.~Teichrieb.
\newblock St-gait++: Leveraging spatio-temporal convolutions for gait-based emotion recognition on videos.
\newblock In {\em Proceedings of the IEEE/CVF Conference on Computer Vision and Pattern Recognition}, pages 302--310, 2024.

\bibitem{liu2024improved}
H.~Liu, C.~Li, Y.~Li, and Y.~J. Lee.
\newblock Improved baselines with visual instruction tuning.
\newblock In {\em Proceedings of the IEEE/CVF Conference on Computer Vision and Pattern Recognition}, pages 26296--26306, 2024.

\bibitem{liu2021swin}
Z.~Liu, Y.~Lin, Y.~Cao, H.~Hu, Y.~Wei, Z.~Zhang, S.~Lin, and B.~Guo.
\newblock Swin transformer: Hierarchical vision transformer using shifted windows.
\newblock In {\em Proceedings of the IEEE/CVF international conference on computer vision}, pages 10012--10022, 2021.

\bibitem{lu2023see}
H.~Lu, X.~Hu, and B.~Hu.
\newblock See your emotion from gait using unlabeled skeleton data.
\newblock In {\em Proceedings of the AAAI Conference on Artificial Intelligence}, volume~37, pages 1826--1834, 2023.

\bibitem{lu2023epic}
H.~Lu, S.~Xu, S.~Zhao, X.~Hu, R.~Ma, and B.~Hu.
\newblock Epic: Emotion perception by spatio-temporal interaction context of gait.
\newblock {\em IEEE Journal of Biomedical and Health Informatics}, 2023.

\bibitem{narayanan2020proxemo}
V.~Narayanan, B.~M. Manoghar, V.~S. Dorbala, D.~Manocha, and A.~Bera.
\newblock Proxemo: Gait-based emotion learning and multi-view proxemic fusion for socially-aware robot navigation.
\newblock In {\em 2020 IEEE/RSJ International Conference on Intelligent Robots and Systems (IROS)}, pages 8200--8207. IEEE, 2020.

\bibitem{omlor2007extraction}
L.~Omlor and M.~A. Giese.
\newblock Extraction of spatio-temporal primitives of emotional body expressions.
\newblock {\em Neurocomputing}, 70(10-12):1938--1942, 2007.

\bibitem{pang2023self}
C.~Pang, X.~Gao, Z.~Chen, and L.~Lyu.
\newblock Self-adaptive graph with nonlocal attention network for skeleton-based action recognition.
\newblock {\em IEEE Transactions on Neural Networks and Learning Systems}, 2023.

\bibitem{paszke2019pytorch}
A.~Paszke.
\newblock Pytorch: An imperative style, high-performance deep learning library.
\newblock {\em arXiv preprint arXiv:1912.01703}, 2019.

\bibitem{peebles2023scalable}
W.~Peebles and S.~Xie.
\newblock Scalable diffusion models with transformers.
\newblock In {\em Proceedings of the IEEE/CVF international conference on computer vision}, pages 4195--4205, 2023.

\bibitem{plizzari2021spatial}
C.~Plizzari, M.~Cannici, and M.~Matteucci.
\newblock Spatial temporal transformer network for skeleton-based action recognition.
\newblock In {\em Pattern recognition. ICPR international workshops and challenges: virtual event, January 10--15, 2021, Proceedings, Part III}, pages 694--701. Springer, 2021.

\bibitem{qu2024llms}
H.~Qu, Y.~Cai, and J.~Liu.
\newblock Llms are good action recognizers.
\newblock In {\em Proceedings of the IEEE/CVF Conference on Computer Vision and Pattern Recognition}, pages 18395--18406, 2024.

\bibitem{randhavane2019identifying}
T.~Randhavane, U.~Bhattacharya, K.~Kapsaskis, K.~Gray, A.~Bera, and D.~Manocha.
\newblock Identifying emotions from walking using affective and deep features.
\newblock {\em arXiv preprint arXiv:1906.11884}, 2019.

\bibitem{randhavane2019learning}
T.~Randhavane, U.~Bhattacharya, K.~Kapsaskis, K.~Gray, A.~Bera, and D.~Manocha.
\newblock Learning perceived emotion using affective and deep features for mental health applications.
\newblock In {\em 2019 IEEE International Symposium on Mixed and Augmented Reality Adjunct (ISMAR-Adjunct)}, pages 395--399. IEEE, 2019.

\bibitem{shi2019two}
L.~Shi, Y.~Zhang, J.~Cheng, and H.~Lu.
\newblock Two-stream adaptive graph convolutional networks for skeleton-based action recognition.
\newblock In {\em Proceedings of the IEEE/CVF conference on computer vision and pattern recognition}, pages 12026--12035, 2019.

\bibitem{vaswani2017attention}
A.~Vaswani, N.~Shazeer, N.~Parmar, J.~Uszkoreit, L.~Jones, A.~N. Gomez, {\L}.~Kaiser, and I.~Polosukhin.
\newblock Attention is all you need.
\newblock {\em Advances in neural information processing systems}, 30, 2017.

\bibitem{venture2014recognizing}
G.~Venture, H.~Kadone, T.~Zhang, J.~Gr{\`e}zes, A.~Berthoz, and H.~Hicheur.
\newblock Recognizing emotions conveyed by human gait.
\newblock {\em International Journal of Social Robotics}, 6:621--632, 2014.

\bibitem{wang20233mformer}
L.~Wang and P.~Koniusz.
\newblock 3mformer: Multi-order multi-mode transformer for skeletal action recognition.
\newblock In {\em Proceedings of the IEEE/CVF Conference on Computer Vision and Pattern Recognition}, pages 5620--5631, 2023.

\bibitem{wang2023efficient}
W.~Wang, J.~Liu, Y.~Su, and W.~Nie.
\newblock Efficient spatio-temporal video grounding with semantic-guided feature decomposition.
\newblock In {\em Proceedings of the 31st ACM International Conference on Multimedia}, pages 4867--4876, 2023.

\bibitem{wasim2024videogrounding}
S.~T. Wasim, M.~Naseer, S.~Khan, M.-H. Yang, and F.~S. Khan.
\newblock Videogrounding-dino: Towards open-vocabulary spatio-temporal video grounding.
\newblock In {\em Proceedings of the IEEE/CVF Conference on Computer Vision and Pattern Recognition}, pages 18909--18918, 2024.

\bibitem{wortman2023hicem}
B.~Wortman and J.~Z. Wang.
\newblock Hicem: A high-coverage emotion model for artificial emotional intelligence.
\newblock {\em IEEE Transactions on Affective Computing}, 15(3):1136--1152, 2023.

\bibitem{wu2024macdiff}
L.~Wu, L.~Lin, J.~Zhang, Y.~Ma, and J.~Liu.
\newblock Macdiff: Unified skeleton modeling with masked conditional diffusion.
\newblock In {\em European Conference on Computer Vision}, pages 110--128. Springer, 2024.

\bibitem{wu2024frequency}
W.~Wu, C.~Zheng, Z.~Yang, C.~Chen, S.~Das, and A.~Lu.
\newblock Frequency guidance matters: Skeletal action recognition by frequency-aware mixed transformer.
\newblock In {\em Proceedings of the 32nd ACM International Conference on Multimedia}, pages 4660--4669, 2024.

\bibitem{xu2022emotion}
S.~Xu, J.~Fang, X.~Hu, E.~Ngai, W.~Wang, Y.~Guo, and V.~C. Leung.
\newblock Emotion recognition from gait analyses: Current research and future directions.
\newblock {\em IEEE Transactions on Computational Social Systems}, 11(1):363--377, 2022.

\bibitem{yin2024msa}
Y.~Yin, L.~Jing, F.~Huang, G.~Yang, and Z.~Wang.
\newblock Msa-gcn: Multiscale adaptive graph convolution network for gait emotion recognition.
\newblock {\em Pattern Recognition}, 147:110117, 2024.

\bibitem{yumeng2024affective}
Z.~YuMeng, L.~Zhen, L.~TingTing, W.~YuanYi, and C.~YanJie.
\newblock Affective-pose gait: perceiving emotions from gaits with body pose and human affective prior knowledge.
\newblock {\em Multimedia Tools and Applications}, 83(2):5327--5350, 2024.

\bibitem{zamir2022restormer}
S.~W. Zamir, A.~Arora, S.~Khan, M.~Hayat, F.~S. Khan, and M.-H. Yang.
\newblock Restormer: Efficient transformer for high-resolution image restoration.
\newblock In {\em Proceedings of the IEEE/CVF conference on computer vision and pattern recognition}, pages 5728--5739, 2022.

\bibitem{zeng2025gaitcycformer}
Q.~Zeng and L.~Shang.
\newblock Gaitcycformer: Leveraging gait cycles and transformers for gait emotion recognition.
\newblock In {\em Proceedings of the AAAI Conference on Artificial Intelligence}, volume~39, pages 9815--9823, 2025.

\bibitem{zhai2024looking}
Y.~Zhai, G.~Jia, Y.-K. Lai, J.~Zhang, J.~Yang, and D.~Tao.
\newblock Looking into gait for perceiving emotions via bilateral posture and movement graph convolutional networks.
\newblock {\em IEEE Transactions on Affective Computing}, 15(3):1634--1648, 2024.

\bibitem{zhang2024hierarchical}
S.~Zhang, J.~Zhang, W.~Song, L.~Yang, and X.~Zhao.
\newblock Hierarchical-attention-based neural network for gait emotion recognition.
\newblock {\em Physica A: Statistical Mechanics and its Applications}, 637:129600, 2024.

\bibitem{zhang2024tt}
T.~Zhang, Y.~Chen, S.~Li, X.~Hu, and C.~P. Chen.
\newblock Tt-gcn: Temporal-tightly graph convolutional network for emotion recognition from gaits.
\newblock {\em IEEE Transactions on Computational Social Systems}, 2024.

\bibitem{zhang2021stst}
Y.~Zhang, B.~Wu, W.~Li, L.~Duan, and C.~Gan.
\newblock Stst: Spatial-temporal specialized transformer for skeleton-based action recognition.
\newblock In {\em Proceedings of the 29th ACM International Conference on Multimedia}, pages 3229--3237, 2021.

\bibitem{zhao2022fine}
S.~Zhao, S.~Yin, H.~Tang, R.~Jin, Y.~Xu, T.~Xu, and E.~Chen.
\newblock Fine-grained micro-expression generation based on thin-plate spline and relative au constraint.
\newblock In {\em Proceedings of the 30th ACM International Conference on Multimedia}, pages 7150--7154, 2022.

\bibitem{zhou2023learning}
H.~Zhou, Q.~Liu, and Y.~Wang.
\newblock Learning discriminative representations for skeleton based action recognition.
\newblock In {\em Proceedings of the IEEE/CVF Conference on Computer Vision and Pattern Recognition}, pages 10608--10617, 2023.

\bibitem{zhou2023fourmer}
M.~Zhou, J.~Huang, C.-L. Guo, and C.~Li.
\newblock Fourmer: An efficient global modeling paradigm for image restoration.
\newblock In {\em International conference on machine learning}, pages 42589--42601. PMLR, 2023.

\bibitem{zhou2022hypergraph}
Y.~Zhou, Z.-Q. Cheng, C.~Li, Y.~Fang, Y.~Geng, X.~Xie, and M.~Keuper.
\newblock Hypergraph transformer for skeleton-based action recognition.
\newblock {\em arXiv preprint arXiv:2211.09590}, 2022.

\bibitem{zhu2022mlst}
X.~Zhu, Y.~Zhou, D.~Wang, W.~Ouyang, and R.~Su.
\newblock Mlst-former: Multi-level spatial-temporal transformer for group activity recognition.
\newblock {\em IEEE Transactions on Circuits and Systems for Video Technology}, 33(7):3383--3397, 2022.

\bibitem{zhu2024temporal}
Z.~Zhu, C.~P. Chen, H.~Liu, and T.~Zhang.
\newblock Temporal group attention network with affective complementary learning for gait emotion recognition.
\newblock In {\em 2024 IEEE International Conference on Bioinformatics and Biomedicine (BIBM)}, pages 3026--3033. IEEE, 2024.

\bibitem{zhuang2020g}
Y.~Zhuang, L.~Lin, R.~Tong, J.~Liu, Y.~Iwamot, and Y.-W. Chen.
\newblock G-gcsn: Global graph convolution shrinkage network for emotion perception from gait.
\newblock In {\em Proceedings of the Asian Conference on Computer Vision}, 2020.

\bibitem{zou2025occluded}
Y.~Zou, N.~He, J.~Sun, X.~Huang, and W.~Wang.
\newblock Occluded gait emotion recognition based on multi-scale suppression graph convolutional network.
\newblock {\em Computers, Materials \& Continua}, 82(1), 2025.

\end{thebibliography}
